\documentclass[conference]{IEEEtran}
\IEEEoverridecommandlockouts

\usepackage{cite}
\usepackage{amsmath,amssymb,amsfonts}
\usepackage{algorithmic}
\usepackage{graphicx}
\usepackage{textcomp}
\usepackage{xcolor}
\usepackage{makecell}
\usepackage{tabularx}
\usepackage{siunitx}
\usepackage{caption}
\usepackage{url}
\usepackage{xcolor}

\def\BibTeX{{\rm B\kern-.05em{\sc i\kern-.025em b}\kern-.08em
    T\kern-.1667em\lower.7ex\hbox{E}\kern-.125emX}}
\begin{document}

\title{Towards a Generalisable Cyber Defence Agent for Real-World Computer Networks}

\author{\IEEEauthorblockN{1\textsuperscript{st} Tim Dudman}
    \IEEEauthorblockA{\textit{Cyber Capability Lead} \\
        \textit{Riskaware}\\
        Bristol, United Kingdom \\
        tim.dudman@riskaware.co.uk}
    \and
    \IEEEauthorblockN{2\textsuperscript{nd} Dr. Martyn Bull}
    \IEEEauthorblockA{\textit{Chief Scientific Officer} \\
        \textit{Riskaware}\\
        Bristol, United Kingdom \\
        martyn.bull@riskaware.co.uk}
}

\maketitle

\begin{abstract}

    Recent advances in deep reinforcement learning for autonomous cyber defence have resulted in agents that can successfully defend simulated computer networks against cyber-attacks. However, many of these agents would need retraining to defend networks with differing topology or size, making them poorly suited to real-world networks where topology and size can vary over time. In this research we introduce a novel set of Topological Extensions for Reinforcement Learning Agents (TERLA) that provide generalisability for the defence of networks with differing topology and size, without the need for retraining. These extensions can be applied to existing autonomous cyber defence agents and may also help address policy learning challenges inherent with large network action spaces.

    Our approach involves the use of heterogeneous graph neural network layers to produce a fixed-size latent embedding representing the observed network state. This representation learning stage is coupled with a reduced, fixed-size, semantically meaningful and interpretable action space. We apply TERLA to a standard deep reinforcement learning Proximal Policy Optimisation (PPO) agent model, and to reduce the sim-to-real gap, conduct our research using Cyber Autonomy Gym for Experimentation (CAGE) Challenge 4. This Cyber Operations Research Gym environment has many of the features of a real-world network, such as realistic Intrusion Detection System (IDS) events and multiple agents defending network segments of differing topology and size. TERLA agents have been designed for host-based defence, and defensive actions are resolved to specific hosts using only observable IDS information that would be available in real-world networks.

    TERLA agents have been evaluated against vanilla PPO agents as well as other approaches including no defence against cyber-attacks and performing random defensive actions. TERLA agents retain the defensive performance of vanilla PPO agents whilst showing improved action efficiency. Generalisability has been demonstrated by showing that all TERLA agents have the same network-agnostic neural network architecture, and by deploying a single TERLA agent multiple times to defend network segments with differing topology and size, showing improved defensive performance and efficiency.

\end{abstract}


\section{Introduction} \label{introduction}

A recent report by the National Cyber Security Centre (NCSC) \cite{ncsc1} provides an assessment of how Artificial Intelligence (AI) will impact the efficacy of cyber operations in the near term. It concludes that AI will almost certainly increase the volume and heighten the impact of cyber-attacks over the next two years through the evolution of adversary Tactics, Techniques and Procedures (TTPs). The NCSC has also called for allies to join forces in the face of a widening gap between escalating cyber threats and the international ability to defend against them \cite{ncsc2}.

To tackle this global threat, in 2024, the Defence Science and Technology Laboratory (Dstl), the Defense Advanced Research Projects Agency (DARPA) and the Canadian Department of National Defence (DRDC) formalised a trilateral collaboration to drive forward AI and cybersecurity systems \cite{army}. This builds on defence-funded research into Autonomous Cyber Defence (ACD) \cite{arcd1, castle} that has been pioneering the development of Deep Reinforcement Learning (DRL) techniques \cite{blackhat} to demonstrate self-defending, self-recovering generation-after-next concepts for military operational platforms. In DRL, agents learn defensive policies through a neural network-based model that observes an environment, selects actions, and improves via rewards within a Markov Decision Process (MDP) \cite{rl}. The stated goal of this research is to achieve fully-autonomous cyber defence, reducing the need for human intervention and ensuring faster response times.

ACD research encompasses both the development of defensive (blue) agents and of the simulated and emulated network environments used for training, evaluating and demonstrating them \cite{arcd2}. Examples of simulated environments include low-fidelity simulations such as YAWNING-TITAN and higher-fidelity simulations such as the Primary-level AI Training Environment (PrimAITE) \cite{environments}. YAWNING-TITAN is a reception-level, abstract, graph based cyber-security simulation environment that supports the training of blue agents against probabilistic adversarial (red) agents, whilst PrimAITE provides a higher-fidelity simulation that includes key network characteristics, background (green) agent pattern of life and more realistic adversary techniques.

Alongside this defence-funded research, The Technical Cooperation Programme (TTCP), a multi-lateral collaboration between the UK, USA, Canada, Australia and New Zealand, has been supporting the development of DRL agents for ACD through the Cyber Autonomy Gym for Experimentation (CAGE) challenges. These public challenges use the Cyber Operations Research Gym (CybORG) environment \cite{cyborg}, with each challenge increasing in fidelity and complexity, culminating in CAGE Challenge 4 (CC4) \cite{cc4} in 2024. This is a Multi-Agent Reinforcement Learning (MARL) challenge where five separate blue agents defend network segments of differing topology and size. The environment includes a high level of realism, with each agent only able to observe the state of its own local network segment, rewards based on both red agent compromise and green agents being able to access critical services during different phases of a mission, and Intrusion Detection System (IDS) alerts that include false positive and false negative events.

\section{Problem Statement} \label{problem-statement}

\begin{figure}[hbtp]
    \centering
    \includegraphics[width=\columnwidth]{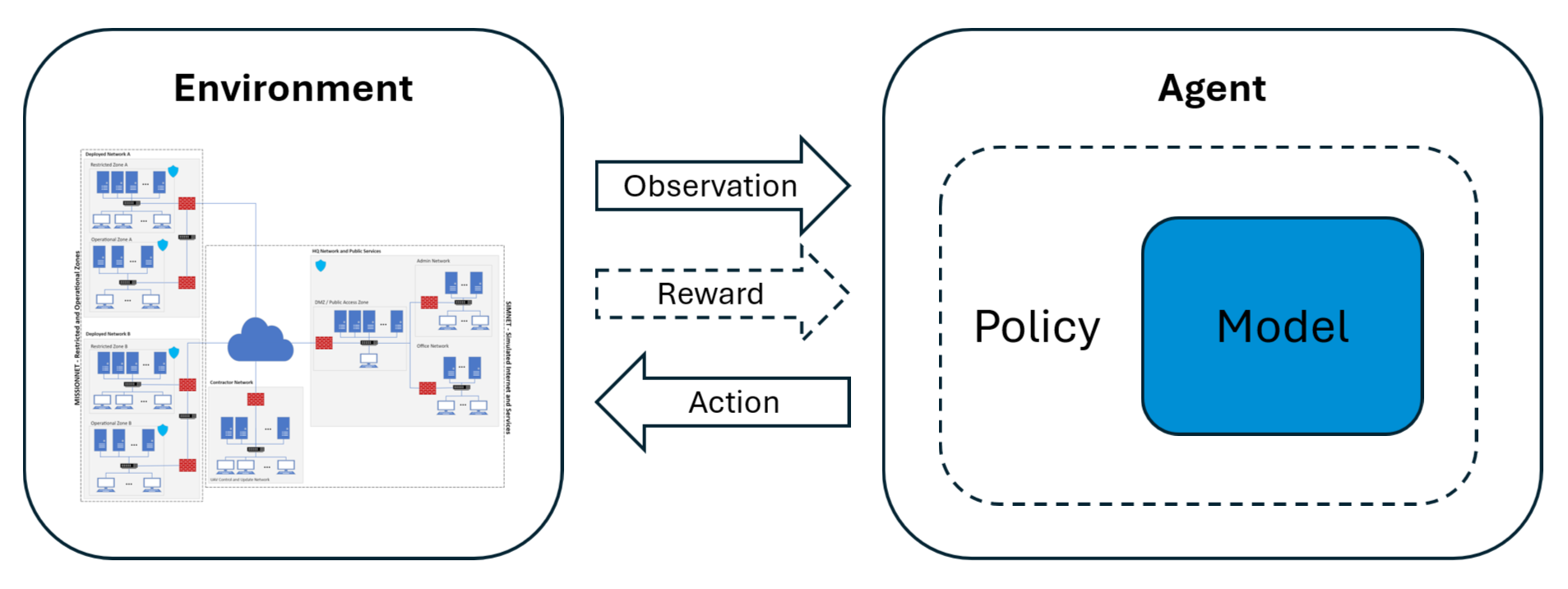}
    \caption{An example of a DRL agent interacting with a cyber environment, exchanging observations, actions and rewards.}
    \label{drl}
\end{figure}

Fig.~\ref{drl} shows a typical DRL agent, composed of a \textit{policy} and a \textit{model}, interacting with a cyber environment by taking actions and receiving observations and rewards\footnote{Note that rewards are used during training and evaluation, but would not be present (or required) in the operational deployment of a pre-trained agent on a real-world network.}. The policy is responsible for training the model (using rewards) and selecting actions based on model output.

The ACD environments discussed in \ref{introduction} are all compatible with Gymnasium, the Application Programming Interface (API) standard for Reinforcement Learning (RL) \cite{gymnasium}. Observations are generated as arrays containing observable features for each subnet and each host in the environment, causing the observation space to be specific to the topology and size of the simulated network. The action space that defines the available actions on each subnet and each host is also specific to the topology and size of the simulated network.

\begin{figure}[hbtp]
    \centering
    \includegraphics[width=\columnwidth]{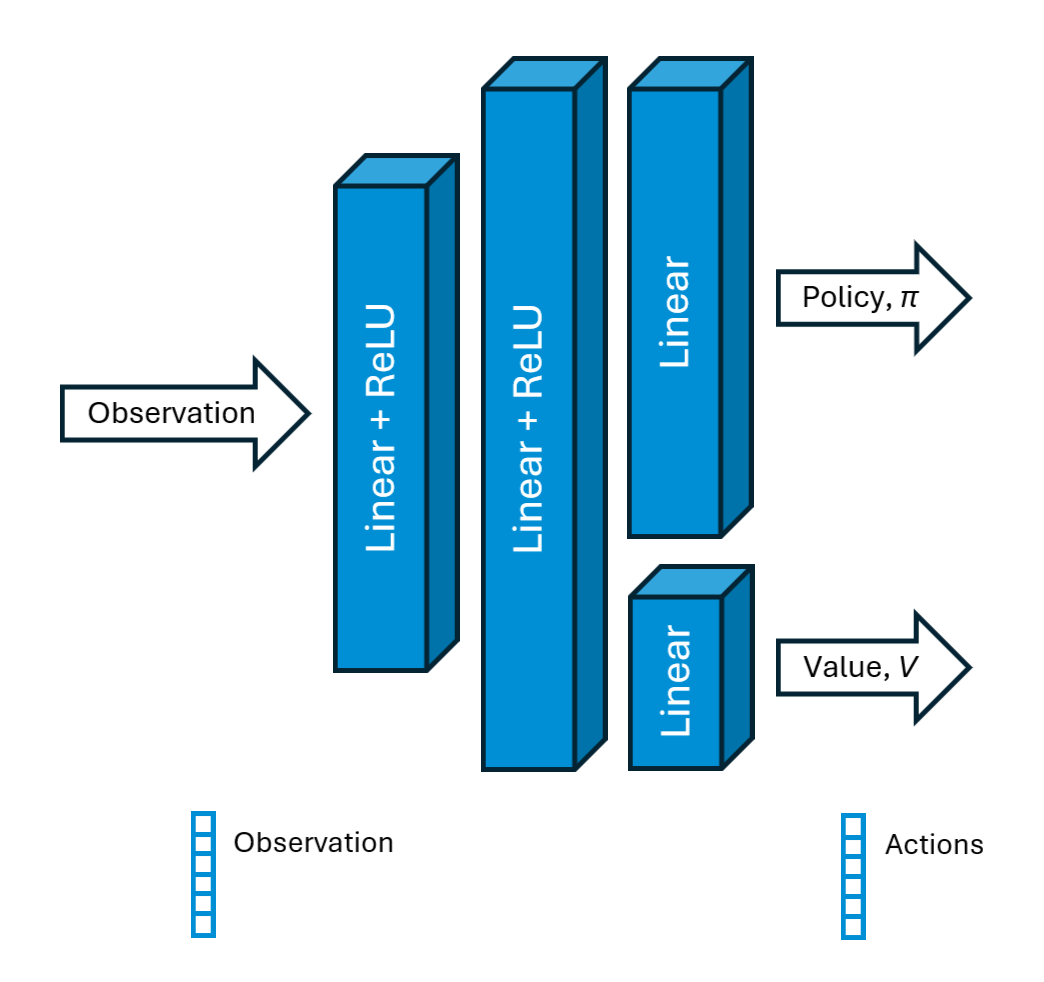}
    \caption{An example of a typical DRL agent model, sized to the observation and action spaces of the environment in which it is trained.}
    \label{ppo-model}
\end{figure}

Fig.~\ref{ppo-model} shows a typical DRL agent model, such as that found in a Proximal Policy Optimisation (PPO) \cite{ppo} agent, consisting of a fully-connected, Multi-Layer Perceptron (MLP) neural network that processes observations to learn a policy, $\pi$, and a value function, $V$. The policy head generates a tensor of logits that corresponds to a probability distribution across the action space.

Because the observation and action spaces are specific to the topology and size of the simulated network, so is the architecture of the neural network (the number of neurons in each layer). This means that a blue agent trained in a simulated network environment can only be deployed to defend a real-world network with identical topology and size. Furthermore, even if that were achieved, the agent would not be able to adapt to dynamic changes such as hosts joining or leaving the network without being retrained with a redefined architecture tailored to the updated network. A related issue is that for large networks, the size of the action space will be very large, increasing training times and providing a significant challenge to learning an optimal policy \cite{challenges}.

Model extensions for state-of-the-art ACD agents, many of which are PPO-based \cite{blackhat}, are therefore proposed, to remove the constraints discussed above without impacting defensive performance and without breaking compatibility with Gymnasium.

\section{Related Work} \label{related-work}

Some research into generalisable DRL for optimising power flow \cite{powerflow}, and indeed ACD \cite{blackhat}, has already been undertaken. This work treated network data as a heterogeneous graph and used non-Gymnasium environments specifically designed to support agents with Graph Neural Network (GNN)-based models\footnote{GNN layers provide convolution operations for graphs and can be used to extract feature embeddings for dimensionality reduction \cite{gnn}.}.

Unlike that research, our solution to the problem set out in \ref{problem-statement} is to develop extensions to existing ACD agents that retain compatibility with Gymnasium. It is inspired by previous research into the Job-Shop Scheduling Problem (JSSP) \cite{jobshop1, jobshop2}. Here, a GNN is used to learn node features that embed the spatial structure of the JSSP (\textit{representation learning}). A standard DRL model is then used to learn an optimum scheduling policy that maps that latent embedding to the best scheduling action (\textit{policy learning}). A PPO-based agent is used to train these two elements in an end-to-end fashion. Unlike the disjunctive graphs used in the JSSP research, computer networks are not fixed-size, so our solution proposes additional pre- and post- policy learning extensions that enhance the concept for ACD, supporting both existing agents and environments.

\section{Methodology}

With the goal of producing DRL model extensions that allow ACD agents to defend real-world networks of different topologies and size without the need for retraining, we base our research on CC4. As discussed in \ref{introduction}, this environment not only allows the training and evaluation of blue agents on network segments with differing topology and size, but it also includes observations that could be expected in a real-world network.

\begin{figure*}[hbtp]
    \centering
    \includegraphics[width=\textwidth]{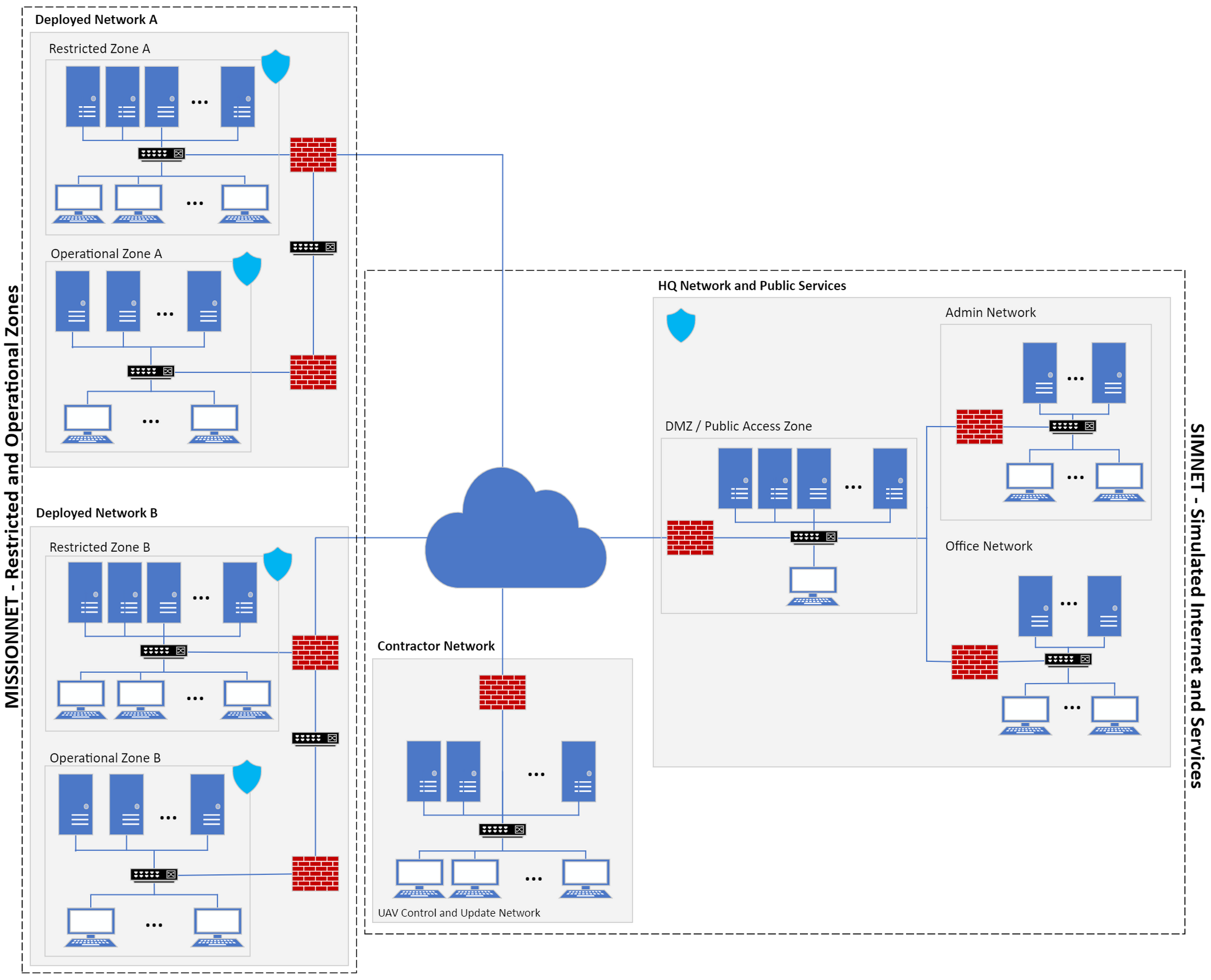}
    \caption{The CC4 network laydown (\textcopyright\ 2024 TTCP) showing the varying topology and size of the five network segments being defended.}
    \label{cc4-network}
\end{figure*}

Fig.~\ref{cc4-network} shows the CC4 network laydown. Two blue agents defend Deployed Network A, one in each security zone. Two more defend Deployed Network B, again, one in each security zone. Each of these agents defends a single subnet comprising 16 hosts. A final blue agent defends three subnets in the HQ and Public Services Network, each again comprising 16 hosts (a total of 48 hosts). Although the observation and action spaces are fixed size for each blue agent, the number of active hosts in each subnet varies between episodes. This means that certain actions in the action space are treated as invalid. The Contractor Network is undefended, and this is where the red agent starts. Other red agents become active in the other subnets once an adversary foothold has been achieved. This allows each compromised subnet to be attacked simultaneously later in an episode. CC4 also includes the concept of mission phases, and the reward function used by CC4 reflects that Operational Zone A is more critical in mission phase 2A, and Operational Zone B is more critical in mission phase 2B.

The CC4 observation space for each agent is detailed in Table~\ref{observation}, and the action space is detailed in Table~\ref{actions}. In this research we are not developing a MARL solution, so the observation message block is not used\footnote{The message block could be included in the graph described in \ref{observation-conversion} to support a MARL solution.}. Combined with the partial observability of network segments, individual agents will not have any information about events observed in subnets they are not defending. We therefore limit our agent to host-based defensive actions using only mission, local network topology and local IDS event observation features that would be observable in the real-world. \textit{Allow traffic zone} and \textit{block traffic zone} actions are not supported, and blocked subnet and communication policy features in the observation space are ignored.

\begin{table}[htbp]
    \caption{The full set of CC4 observation features available to blue agents, with vector sizes specific to the topology and size of the simulated network.}
    \begin{center}
        \begin{tabularx}{\linewidth}{|X|X|}
            \hline
            \textbf{Description} & \textbf{Value} \\
            \hline
            Mission phase & \makecell[l]{0: Mission Phase 1 \\ 1: Mission Phase 2A \\ 2: Mission Phase 2B} \\
            \hline
            Subnet vector & \makecell[l]{1: Current subnet \\ 0: All Other subnets} \\
            \hline
            Blocked subnets vector & \makecell[l]{1: Subnet blocked \\ 0: Subnet not blocked} \\
            \hline
            Communication policy vector & \makecell[l]{1: Subnet should be blocked \\ 0: Subnet should not be blocked} \\
            \hline
            \raggedright Host malicious process event vector & \makecell[l]{1: Malicious process detected               \\ 0: No events} \\
            \hline
            \raggedright Host malicious network event vector & \makecell[l]{1: Malicious connection detected            \\ 0: No events} \\
            \hline
            Message block & Optional 8-bit agent-specific messages from other agents \\
            \hline
        \end{tabularx}
        \label{observation}
    \end{center}
\end{table}

\begin{table}[htbp]
    \caption{The full set of CC4 actions available to blue agents, applicable to each subnet and host being defended.}
    \begin{center}
        \begin{tabularx}{\linewidth}{|p{0.2\linewidth}|X|p{0.2\linewidth}|}
            \hline
            \textbf{Action type} & \textbf{Description} & \centering \arraybackslash \textbf{Duration (timesteps)} \\
            \hline
            Sleep & Take no action this timestep. & \centering \arraybackslash 1 \\
            \hline
            Monitor & Collect information about flagged malicious activity on the network. This action occurs automatically as a default action. & \centering \arraybackslash 1 \\
            \hline
            Analyse & Collect further information relating to malware on a specific host to enable the blue agent to better identify if the red agent is present on the system. & \centering \arraybackslash 2 \\
            \hline
            Remove & Attempt to remove the red agent from a host by destroying malicious processes, files and services. & \centering \arraybackslash 3 \\
            \hline
            Restore & Restore a system to a known good state. This has significant consequences for system availability. & \centering \arraybackslash 5 \\
            \hline
            Deploy Decoy & Setup a decoy service on a specified host. When a red agent discovers or exploits a decoy service, the blue agent will receive alerts involving that host or service. & \centering \arraybackslash 2 \\
            \hline
            Allow Traffic & Allow traffic to and from the specified zone. & \centering \arraybackslash 1 \\
            \hline
            Block Traffic & Block traffic to and from the specified zone. If green agents are attempting to communicate to that zone this will result in penalties. & \centering \arraybackslash 1 \\
            \hline
        \end{tabularx}
        \label{actions}
    \end{center}
\end{table}

\begin{figure*}[hbtp]
    \centering
    \includegraphics[width=\textwidth]{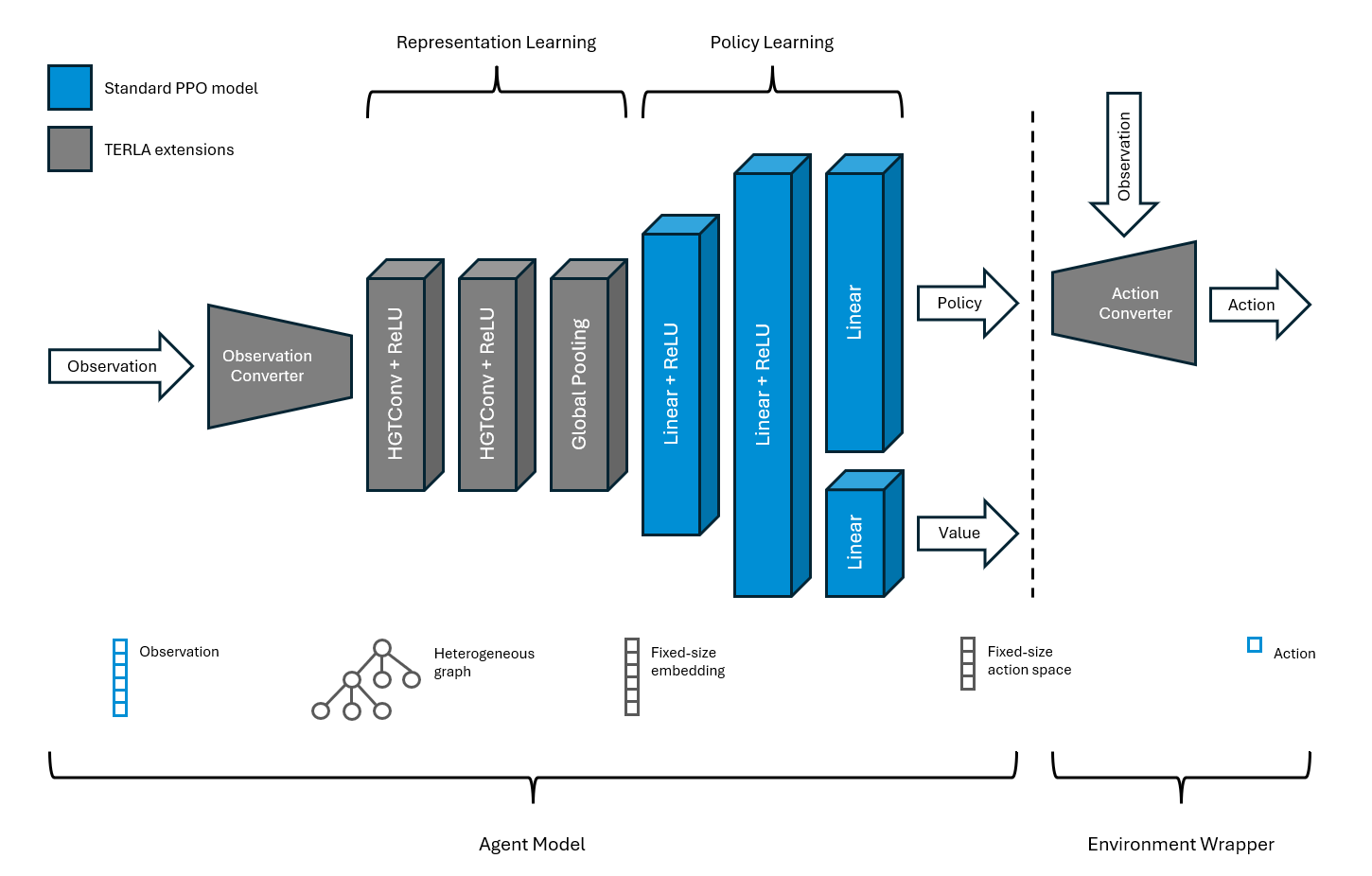}
    \caption{The TERLA architecture applied to a standard PPO model.}
    \label{terla-model}
\end{figure*}

CC4 is integrated with RLlib \cite{rllib}, a scalable open source library for RL that supports production-level MARL workloads. We therefore base our agent on the RLlib implementation of PPO and introduce a novel set of Topological Extensions for Reinforcement Learning Agents (TERLA) that extends the existing RLlib PPO model. The TERLA architecture is shown in Fig.~\ref{terla-model}, and consists of four key elements:

\begin{itemize}
    \item Initial conversion of the observation into a heterogeneous graph.
    \item Representation learning of a fixed-size latent embedding using an encoder.
    \item A policy that learns to select actions from a reduced, fixed-size, semantically meaningful and interpretable action space.
    \item Final conversion of TERLA actions back to the environment action space using only information available in the original observation.
\end{itemize}

These elements are discussed in \ref{observation-conversion}, \ref{representation-learning}, \ref{policy-learning} and \ref{action-conversion} below. \ref{additional-techniques} discusses two additional techniques that were implemented to support TERLA agents in CC4.

\subsection{Observation Conversion} \label{observation-conversion}

\begin{figure*}[hbtp]
    \centering
    \includegraphics[width=\textwidth]{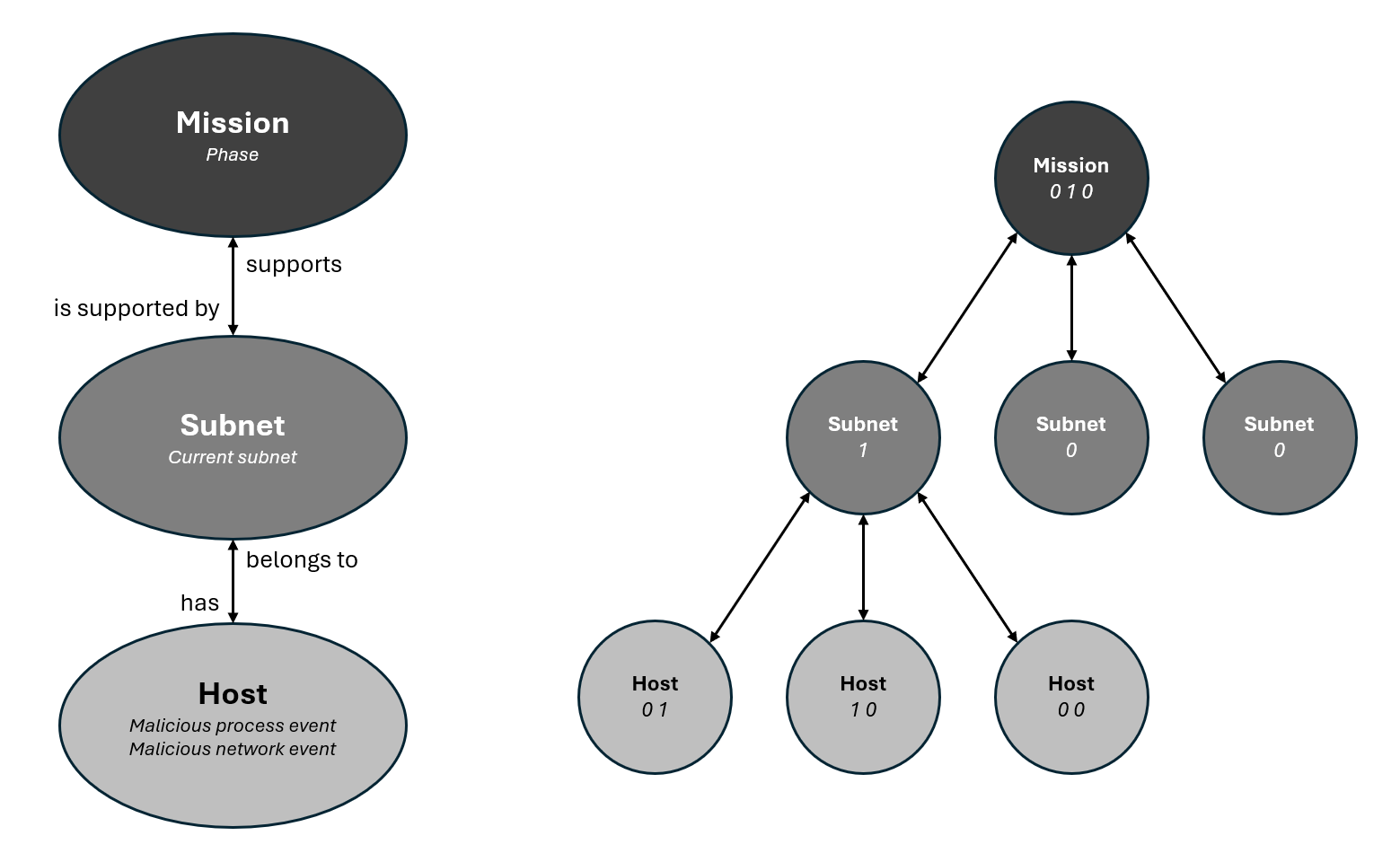}
    \caption{The TERLA heterogeneous graph (schema on the left and an example instance as seen by an agent on the right).}
    \label{heterogeneous-graph}
\end{figure*}

Similar to prior solutions to the JSSP discussed in \ref{related-work}, TERLA agents learn node features that embed the spatial structure of the observed environment state. To facilitate this, each observation from the environment is converted into the heterogeneous graph shown in Fig.~\ref{heterogeneous-graph}. This captures the key features from the environment identified above\footnote{Note that mission phase is now one-hot encoded.}. Because the graph is undirected, the encoder is not limited in the relationships it can learn. In order to adhere to the RLlib API, which will only send batches of tensors to the agent, the conversion happens in the agent model\footnote{This can be inefficient because by the time the observation reaches the agent, RLlib has already moved it to the Graphics Processing Unit (GPU) for processing by the model. When the observation is converted, it is first moved to the Central Processing Unit (CPU), before being converted to a heterogeneous graph and moved back to the GPU. It should be noted that this inefficiency is specific to the use of RLlib and is not inherent in the design of TERLA.}.

\subsection{Representation Learning} \label{representation-learning}

The encoder comprises two Heterogeneous Graph Transformer (HGT) \cite{hgt} layers, each with a Rectified Linear Unit (ReLU) activation function and a global pooling layer that sums host features across the node dimension. It generates a fixed-size, reduced-dimensionality latent embedding representing the observed network state. This is used, after normalisation, by the standard PPO model for downstream policy learning. Both the representation learning and policy learning stages are jointly trained, enabling task-aligned feature extraction in the latent space and improved generalisation to changes in network topology, guided by the reward signal.

HGT layers were chosen for their compatibility with the heterogeneous graphs used to model CC4 observations. They also provide scalability and employ a type-specific attention mechanism to capture individual relationships within the data. The initial encoder layer is sized to the input graph schema, and the subsequent encoder hidden layer is sized using \eqref{gnn-size}.

\begin{equation}
    \begin{aligned}
        \textit{Encoder hidden layer size} &= \bigl( \textit{number of host features} \\
        &\quad + \textit{number of TERLA actions} \bigr) \\
        &\quad \times 10
    \end{aligned}
    \label{gnn-size}
\end{equation}

This gives an encoder hidden layer size of 70 neurons (2 host features and 5 TERLA actions, see \ref{observation-conversion} and \ref{policy-learning}), corresponding to a latent embedding of 70 dimensions that provides a constant input dimension for downstream policy learning. The hidden layers in the standard PPO model are set to twice the size of the encoder hidden layer (140 neurons). This provides topology and size invariance across different networks, given the graph schema described in \ref{observation-conversion}. Larger hidden layer sizes were not found to produce better results in CC4, and the approach of sizing hidden layers based on the graph schema limits the size of the neural network, which reduces the scale of the policy learning challenge, especially for large networks. Furthermore, without the need to manually set the hidden size hyperparameter, TERLA agents should be simpler to tune.

\subsection{Policy Learning} \label{policy-learning}

The architecture of the standard PPO model remains unchanged (apart from the layer size now being fixed), but the policy being learned selects between actions in a reduced, fixed-size, semantically meaningful and interpretable action space where the action is targeted after selection, based on the action description. The action descriptions are formulated so that the target of the action can be resolved solely from IDS information in the original observation, as would be available in a real-world network. Table~\ref{terla-action-mapping} details the TERLA action space for each agent, and how it maps to the CC4 action space.

\begin{table}[htbp]
    \caption{The mapping between TERLA actions and CC4 actions.}
    \begin{center}
        \begin{tabularx}{\linewidth}{|X|X|}
            \hline
            \textbf{TERLA action} & \textbf{CC4 action} \\
            \hline
            Do nothing & Sleep \\
            \hline
            Check for malware on the least compromised host & Analyse (subnet X, host Y) \\
            \hline
            Remove malicious user shell from the most compromised host & Remove (subnet X, host Y) \\
            \hline
            Re-image the most compromised host & Restore (subnet X, host Y) \\
            \hline
            Deploy honeypot service on the most compromised host & Deploy Decoy (subnet X, host Y) \\
            \hline
        \end{tabularx}
        \label{terla-action-mapping}
    \end{center}
\end{table}

\subsection{Action Conversion} \label{action-conversion}

TERLA actions are targeted after selection by finding the first instance of a host in the network segment being defended that meets the action description\footnote{This is because in CC4 there is no information in the observation that would allow target prioritisation. In an environment where this information was available, more advanced targeting could be implemented.}. This is either the most compromised or least compromised host. Host compromise is determined by analysis of the host malicious process and network event features in the observation. Hosts with both events detected are considered the most compromised. Hosts with only malicious process events detected are the next most compromised, followed by hosts with only malicious network events detected. Determination of least compromised hosts follows the inverse logic, starting with hosts with no events detected. Through action targeting, invalid actions, and therefore any associated negative rewards, are avoided. Unlike observation conversion, action conversion takes place in an \textit{environment wrapper}. This is a mechanism provided by CC4 to allow environment customisation\footnote{For efficiency, observation conversion could also take place here on the CPU, with the unpacked heterogeneous graph being passed to the agent so that RLlib can move it to the GPU before packing.}.

\subsection{Additional techniques} \label{additional-techniques}

Because TERLA agents learn a policy from a fixed-size latent embedding, two complexities of CC4 needed to be addressed to facilitate successful learning. The first of these concerns action duration. In CC4, blue agents are not permitted to perform new actions while existing actions are in progress. In the RL MDP supported by RLlib, observations and actions pass between the environment and agents every timestep, so CC4 will silently convert any new actions from an agent with an ongoing action to the \textit{sleep} action. The rewards an agent sees do not therefore necessarily reflect the action that was requested. To stop agents learning from this false experience, a technique we have called \textit{action waiting} was implemented in the environment wrapper. This technique stops observations being sent to agents that have ongoing actions. Rewards continue to be sent as normal, and this is important during evaluation.

The second complexity concerns the fact that CC4 returns shared rewards to all agents. This means that each agent receives the same reward on a given timestep related to the overall health in the entire network. Therefore individual agents will receive rewards that are not just related to the health of the network segment they defend, but all network segments. TERLA agents need more specific rewards to train successfully, so \textit{reward shaping} was also implemented in the environment wrapper. Shaped rewards are based solely on the difference in network segment health from the previous timestep and are calculated for each agent based only on the network segment they defend. Network health is determined by the number of red agent sessions and the health of green services, taking into account Operational Technology (OT) using \eqref{network-health}.

\begin{equation}
    \begin{aligned}
        \textit{reward} &= - \sum_{i=1}^{h} \biggl( \textit{number of red sessions}_i \\
        &\quad + \sum_{j=1}^{s} \bigl( \textit{service unreliability}_{ij} \times \textit{OT multiplier}_i \bigr) \biggr)
    \end{aligned}
    \label{network-health}
\end{equation}

where \textit{h} is the number of hosts in the network segment being defended, \textit{s} is the number of green sessions on each host, and \textit{service unreliability} is between 0 (active and totally reliable) and 1 (inactive or totally unreliable). The \textit{OT multiplier} is set at 1 for standard hosts and 2 for OT hosts.

Unlike the original CC4 shared rewards, shaped rewards do not explicitly take the ability of green agents to access services into account\footnote{This is partly because green agent behaviour is simulated by CybORG and not accessible in the environment wrapper.}. They are also unrelated to mission phase, and there are no explicit penalties for using the \textit{restore} action. Although simpler, shaped rewards allow TERLA agents to learn to maintain service availability and defend OT hosts across the different mission phases. The agent still observes mission phase, so it can learn different strategies as a mission progresses (see \ref{evaluation}), but these will be based purely on observed mission phase and IDS events, not rewards. Because the original CC4 shared rewards are used during evaluation, defensive performance is still evaluated against rewards that do take the ability of green agents to access services in different mission phases into account.

\section{Results}

TERLA agents have been trained and evaluated in CC4 to demonstrate generalisability without any loss of defensive performance compared to vanilla PPO agents. Two different deployments have been investigated\textemdash separate TERLA agents deployed in each network segment (the standard CC4 setup), and a single TERLA agent deployed multiple times to defend each network segment. This second deployment is only possible because TERLA agents are agnostic to network topology and size\footnote{CC4 observation padding was enabled to facilitate the training of a single TERLA agent in each network segment because RLlib requires that all observations are the same size for batching. However, the TERLA agent model neither requires nor uses the padding, and padding was not enabled during evaluation.}. These TERLA agents have been evaluated alongside vanilla PPO agents, and PPO agents that use action waiting and reward shaping to show that those techniques on their own do not lead to improved defensive performance. Additionally, evaluation results for \textit{sleep} and \textit{random} agents have been included for context. These agents simulate taking no defensive actions and performing random defensive actions respectively. All ACD agents should of course perform better than these baselines.

\subsection{Training} 

Table~\ref{hyperparameters} lists the hyperparameters that were found to give effective performance for vanilla PPO agents in CC4. These hyperparameters also produced TERLA agents with effective performance, so were used in all training runs to allow a fair comparison. Hidden layer size is only relevant to PPO agents, but it is worth noting that TERLA agent models comprise fewer neurons overall (see \ref{representation-learning}), despite having additional encoder layers.

\begin{table}[htbp]
    \caption{The hyperparameters used to train all agents.}
    \begin{center}
        \begin{tabularx}{\linewidth}{|X|X|}
            \hline
            \textbf{Hyperparameter} & \centering \arraybackslash \textbf{Value} \\
            \hline
            Bellman discount factor ($\gamma$) & \centering \arraybackslash 0.97 \\
            \hline
            Learning rate ($\alpha$) & \centering \arraybackslash \SI{1.0e-4}{} \\
            \hline
            Entropy coefficient ($\beta$) & \centering \arraybackslash 0.01 \\
            \hline
            Hidden layer size (PPO only) & \centering \arraybackslash 256, 256 \\
            \hline
            Rollout fragment length & \centering \arraybackslash 128 \\
            \hline
            Training batch size & \centering \arraybackslash 2,048 \\
            \hline
            Episode length & \centering \arraybackslash 500 \\
            \hline
            Iterations & \centering \arraybackslash 500 \\
            \hline
        \end{tabularx}
        \label{hyperparameters}
    \end{center}
\end{table}

\begin{figure*}[hbtp]
    \centering
    \includegraphics[width=\textwidth]{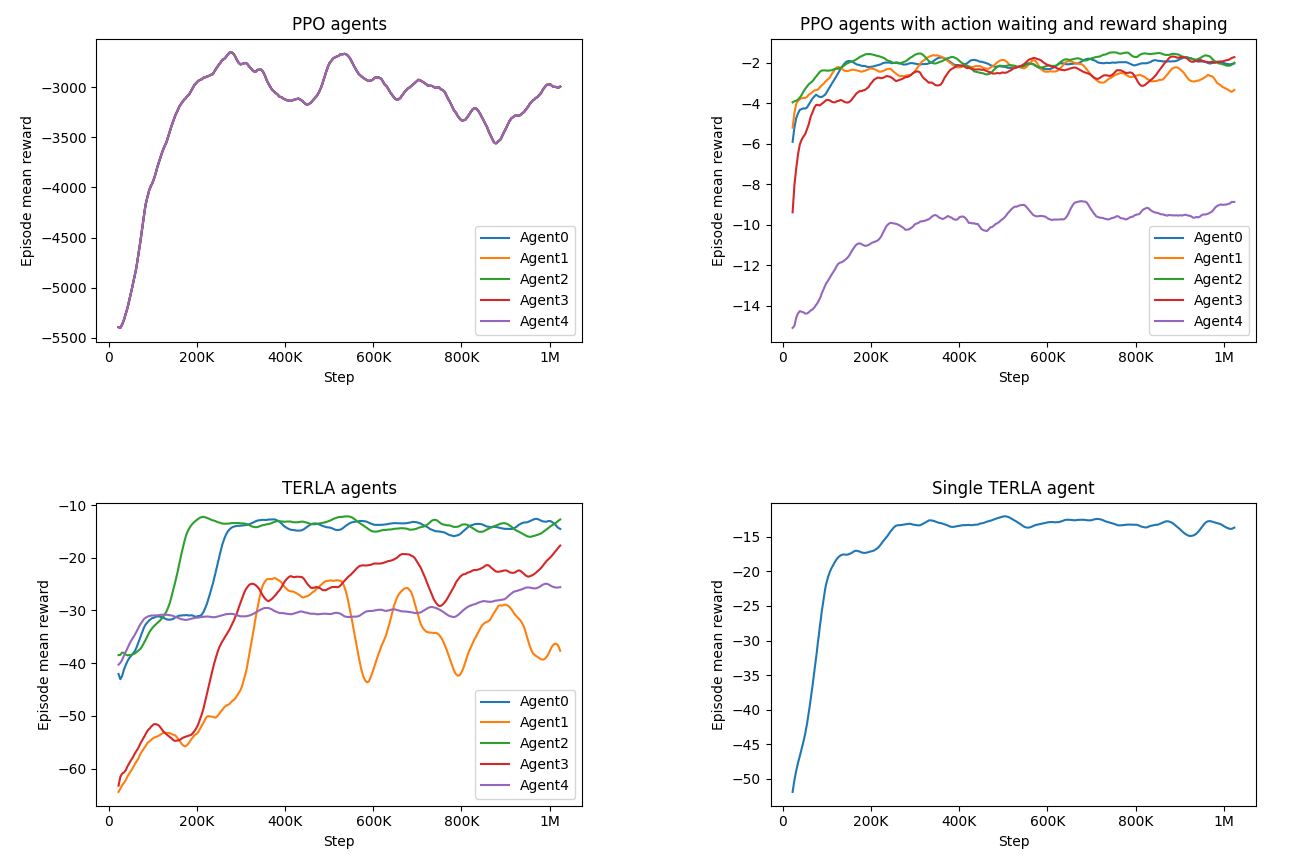}
    \caption{Training plots for all agent types showing convergence to a defensive policy.}
    \label{training}
\end{figure*}

Fig.~\ref{training} shows training plots for each type of agent (sleep and random agents are not included as these do not require training). Vanilla PPO agents train using the original CC4 shared rewards, where each agent receives the same reward. These agents converge on a policy in around 300K steps, but training for all agents was conducted out to 500 iterations, which, given a training batch size of 2,048, equates to approximately 1M steps or 2,000 episodes of CC4 (at 500 steps per episode).

PPO agents with action waiting and reward shaping, separate TERLA agents, and the single TERLA agent all use shaped rewards. It can be seen that some separate TERLA agents require the full amount of training iterations to converge on a policy and have benefitted from the exploration encouraged by a non-zero entropy coefficient. PPO agents with action waiting and reward shaping converge to a higher episode mean reward, although it can be seen that \textit{Agent4} (the agent defending the larger HQ and Public Services Network) did not achieve the same episode mean reward as the agents defending the smaller network segments. Like vanilla PPO agents, the single TERLA agent also converges on a policy in around 300K steps. This is because it rapidly gains experience across each network segment as it trains. The training plots show that all agents converge to a defensive policy, but deterministic evaluation of each trained agent against the original CC4 shared rewards, and against baseline sleep and random agents is required to assess defensive performance in CC4.

\subsection{Evaluation} \label{evaluation}

\begin{figure*}[http]
    \centering
    \includegraphics[width=\textwidth]{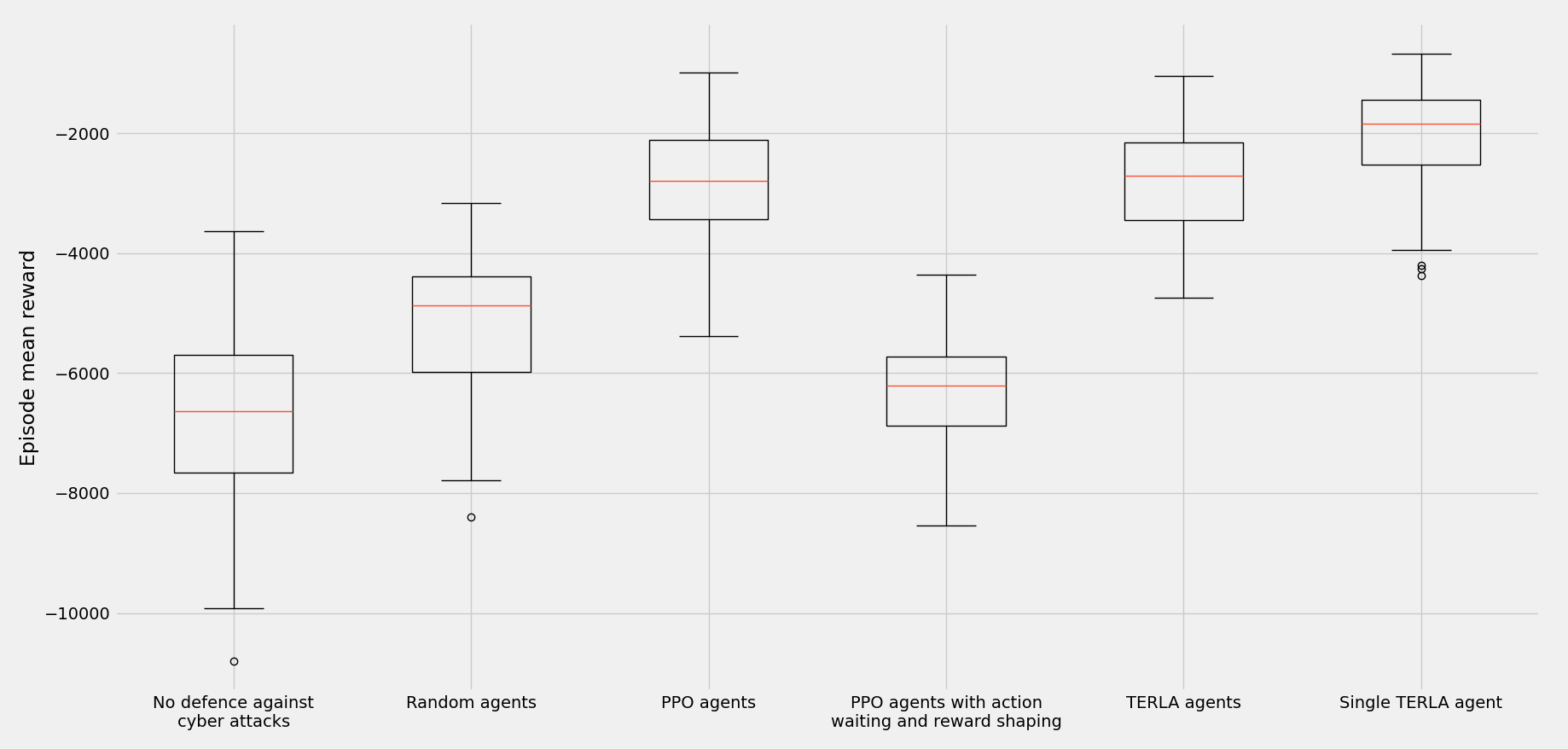}
    \caption{Evaluation results showing that separate TERLA agents retain the defensive performance of PPO agents, and that the single TERLA agent shows improved performance.}
    \label{evaluation-results}
\end{figure*}

Fig.~\ref{evaluation-results} shows the relative defensive performance of each type of agent as a box-and-whisker plot. Evaluation was conducted over 100 episodes, each with an episode length of 500 steps (the same criteria used for CC4 competitive submissions)\footnote{It should be noted that the best performing CC4 competitive submissions employed heuristics or tracked observations and actions in a MARL solution \cite{cc4}. These agents were designed specifically to win the challenge, hence achieving a higher episode mean reward than vanilla PPO agents.}. Sleep agents provide a baseline performance where no defensive actions are taken. Random agents have limited defensive performance but occasionally perform useful actions. As expected, vanilla PPO agents are able to learn a policy that achieves a higher episode mean reward than either of these baselines. However, PPO agents with action waiting and reward shaping performed poorly, since without TERLA action targeting they cannot exploit shaped rewards effectively. It is likely that with such a large action space, shaped rewards did not capture enough of the CC4 objectives, particularly those around maintaining green agent access to services, to allow these agents to learn an effective policy. This result shows that action waiting and reward shaping by itself does not improve defensive performance. Separate TERLA agents retain the defensive performance of vanilla PPO agents, and the single TERLA agent showed improved defensive performance. Table~\ref{numeric-results} records the episode mean reward, standard deviation and relative defensive effectiveness (compared to the sleep agent) of all agents, ranked in order of ascending effectiveness.

\begin{figure*}[hbtp]
    \centering
    \includegraphics[width=\textwidth]{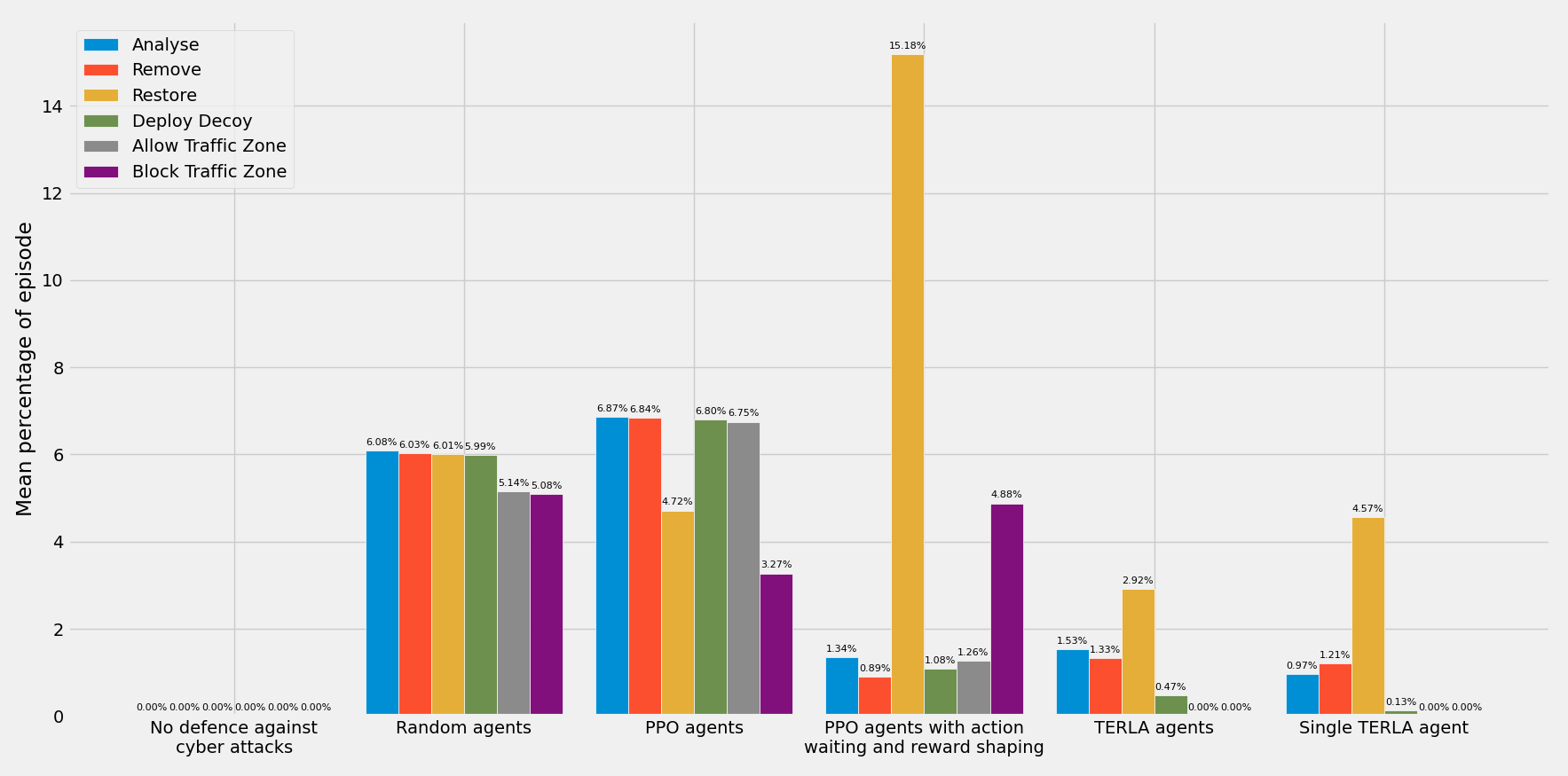}
    \caption{Action distributions for each type of agent, showing that TERLA agents are more efficient than PPO agents.}
    \label{action-distribution}
\end{figure*}

Fig.~\ref{action-distribution} shows the distribution of actions taken by each type of agent, averaged across all evaluation episodes. \textit{Sleep} and \textit{monitor} actions are not shown as these are effectively non-action. Sleep agents take no actions as expected. Random agents randomly select actions, although fewer \textit{allow traffic zone} and \textit{block traffic zone} actions due to there being fewer of these actions in the action space. Vanilla PPO agents learn to restore and block less because these actions disrupt green activity (and there are explicit penalties for \textit{restore} actions). PPO agents with action waiting and reward shaping learn to do the opposite because they lack the ability to target actions effectively using shaped rewards. TERLA agents learn to target \textit{restore} actions effectively, performing more of them than any other action, but fewer overall than other agents. TERLA agents (separate and single) only perform actions between 6\% and 7\% of the time, remaining dormant at all other times. This makes TERLA agents more efficient than vanilla PPO agents, which perform actions around a third of the time.

\begin{table}[htbp]
    \caption{Evaluation results comparing the relative defensive effectiveness of all agent types.}
    \begin{center}
        \begin{tabularx}{\linewidth}{|X|p{0.2\linewidth}|p{0.2\linewidth}|p{0.2\linewidth}|}
            \hline
            \textbf{Agent} & \centering \textbf{Episode mean reward} & \centering \textbf{Standard deviation} & \centering \arraybackslash \textbf{Relative defensive effectiveness} \\
            \hline
            Sleep agents & \centering -6650 & \centering 1420 & \centering \arraybackslash 0\% \\
            \hline
            PPO agents with action waiting and reward shaping & \centering -6300 & \centering 940 & \centering \arraybackslash 5\% \\            \hline
            Random agents & \centering -5150 & \centering 1049 & \centering \arraybackslash 23\% \\
            \hline
            PPO agents & \centering -2825 & \centering 912 & \centering \arraybackslash 58\% \\
            \hline
            TERLA agents & \centering -2773 & \centering 917 & \centering \arraybackslash 58\% \\
            \hline
            Single TERLA agent & \centering -2048 & \centering 799 & \centering \arraybackslash 69\% \\
            \hline
        \end{tabularx}
        \label{numeric-results}
    \end{center}
\end{table}

\begin{figure*}[hbtp]
    \centering
    \includegraphics[width=\textwidth]{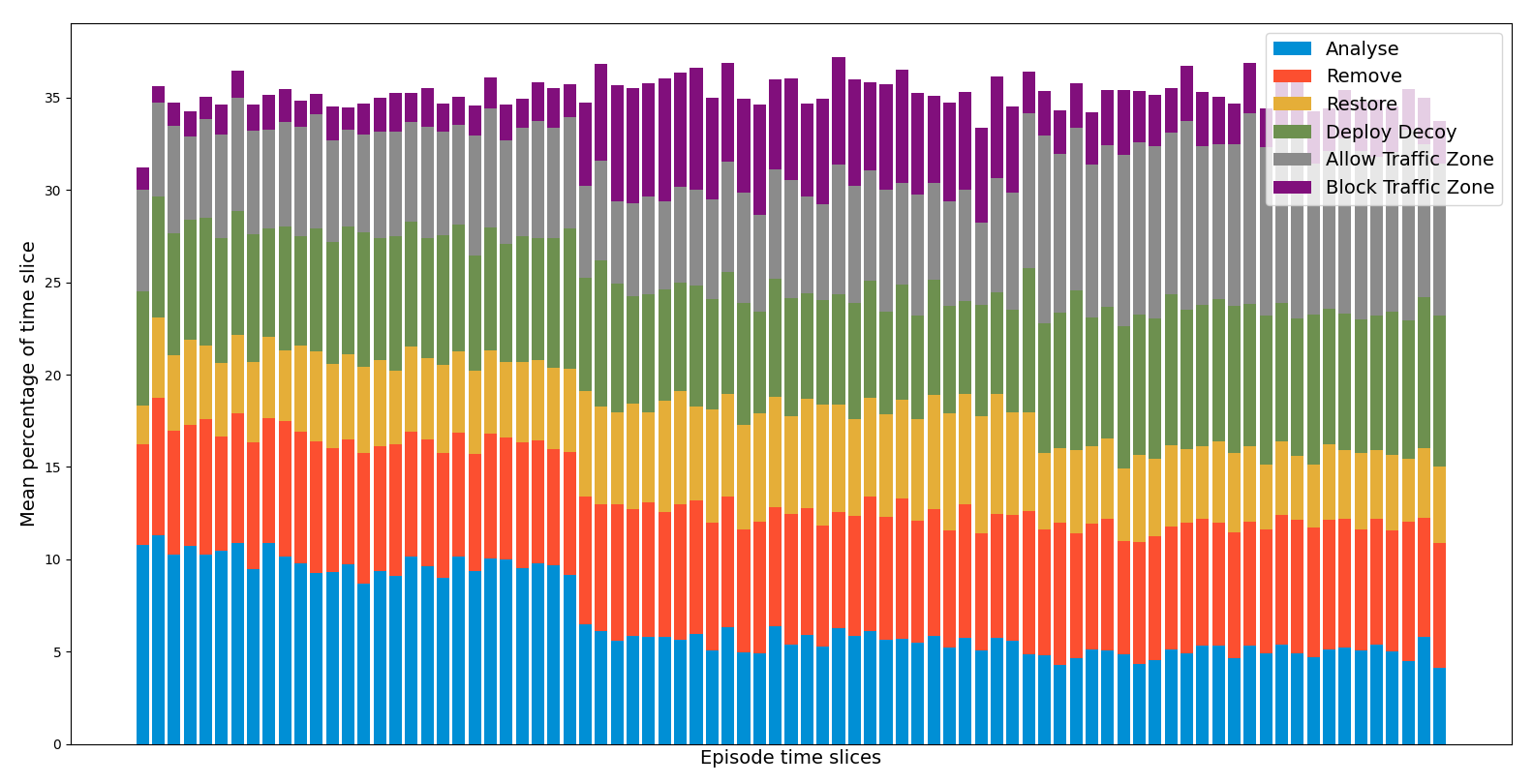}
    \caption{The actions PPO agents took as episodes progressed, showing relatively uniform behaviour across mission phases.}
    \label{ppo-actions}
\end{figure*}

\begin{figure*}[hbtp]
    \centering
    \includegraphics[width=\textwidth]{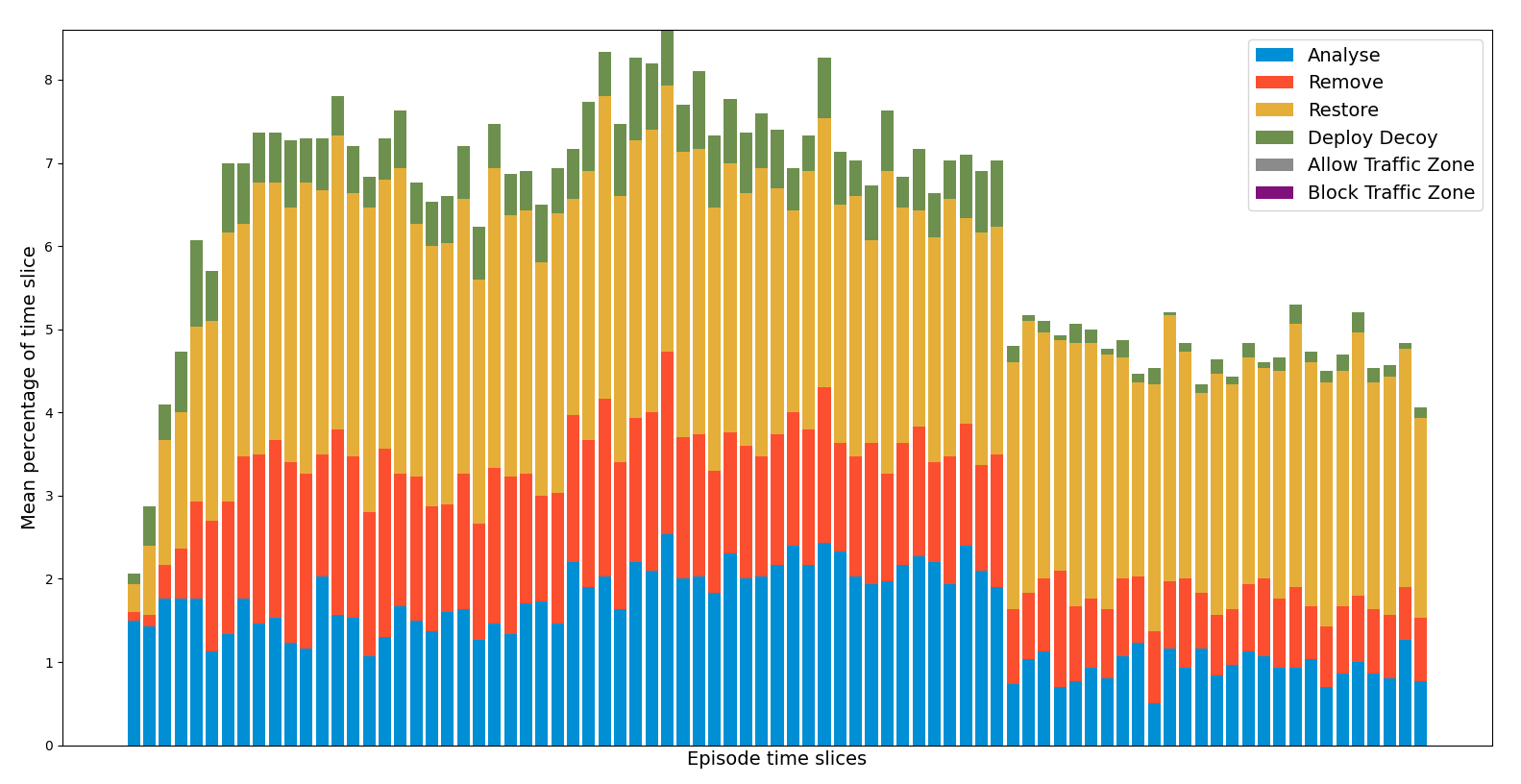}
    \caption{TERLA agents adapt actions across mission phases, with fewer \textit{analyse} and \textit{remove} actions in Phase 2B.}
    \label{terla-actions}
\end{figure*}

\begin{figure*}[hbtp]
    \centering
    \includegraphics[width=\textwidth]{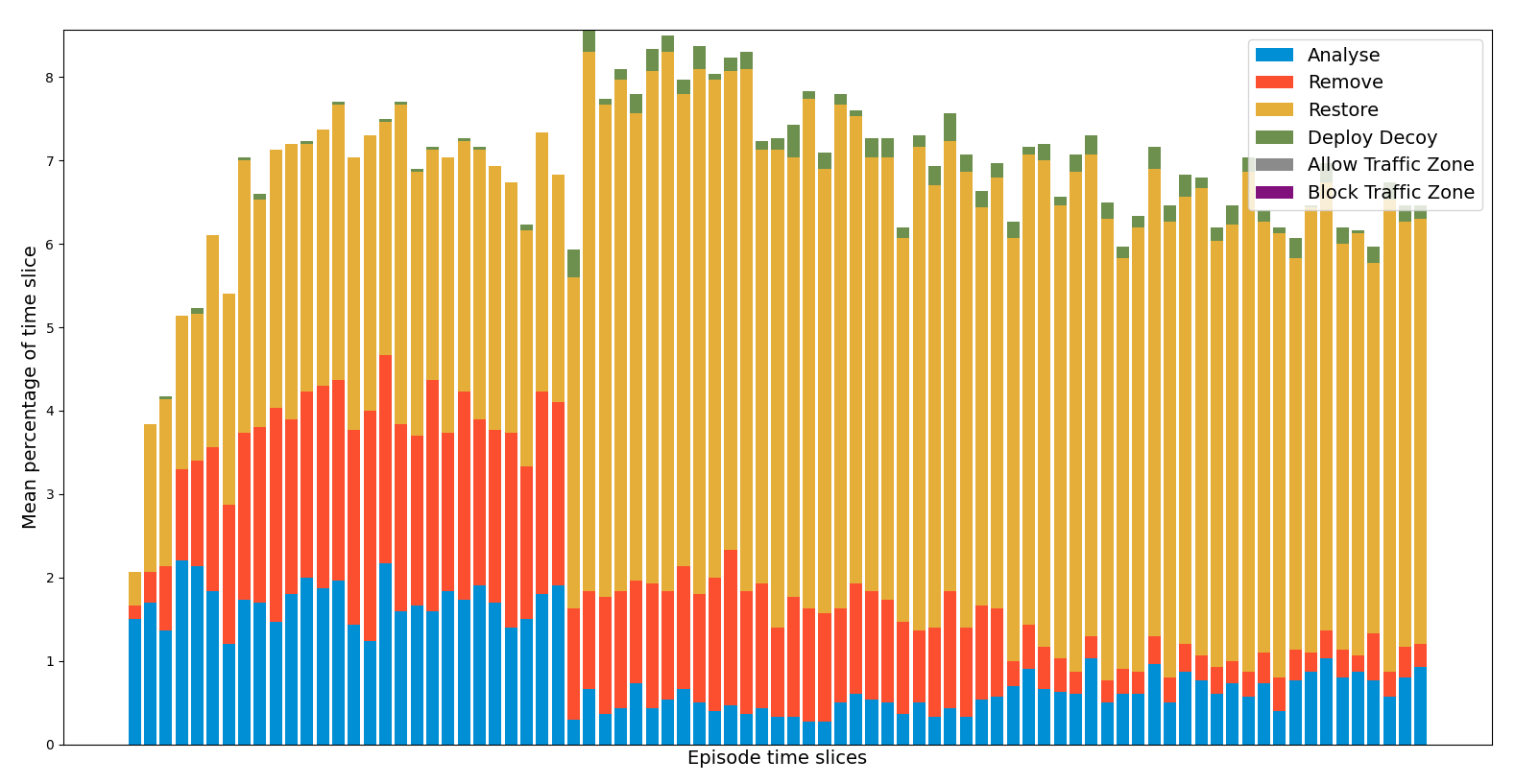}
    \caption{The single TERLA agent reduces \textit{analyse} and \textit{remove} actions, and increases \textit{restore} actions from Phase 2A onward.}
    \label{single-terla-actions}
\end{figure*}

Fig.~\ref{ppo-actions}, Fig.~\ref{terla-actions} and Fig.~\ref{single-terla-actions} show the actions that vanilla PPO agents, separate TERLA agents and the single TERLA agent take over the course of an episode (averaged across all evaluation episodes). It can be seen that TERLA agents exhibit larger differences in behaviour in different mission phases (each mission phase lasting one third of an episode) than vanilla PPO agents, despite not having rewards that reflect mission phase. Both TERLA agent deployments perform fewer \textit{analyse} and \textit{remove} actions in later mission phases, with the single TERLA agent learning to perform fewer of these actions (and more \textit{restore} actions) from mission phase 2A onwards. 

\section{Conclusion}

In this paper, we have proposed TERLA\textemdash extensions to existing ACD agents that provide generalisability for the defence of computer networks with differing topology and size, without the need for retraining. Unlike prior GNN approaches to ACD generalisability, TERLA maintains compatibility with existing environments and may also help address policy learning challenges inherent with large network action spaces.

The TERLA model architecture converts network observations into a heterogeneous graph, and a representation learning stage produces a fixed-size latent embedding representing the observed network state. This embedding is independent of network topology and size and allows existing policy learning elements to learn to select actions from a reduced, semantically meaningful and interpretable action space where the action is targeted after selection using only IDS information in the original observation. Action waiting and reward shaping techniques have also been developed to overcome complexities associated with the use of CC4.

TERLA agents have been designed for host-based defence and have been trained and evaluated in CC4 to demonstrate generalisability without any loss of defensive performance compared to vanilla PPO agents. No defence against cyber-attacks, performing random defensive actions, and PPO agents with action waiting and reward shaping were also evaluated to provide context to the results. All TERLA agents have the same internal neural network architecture, and this allowed two different deployments to be demonstrated\textemdash separate TERLA agents deployed in each network segment, and the same TERLA agent deployed multiple times to defend each network segment. The latter of these demonstrated generalisability for the defence of networks with differing topology and size, and showed improved defensive performance compared to vanilla PPO agents. TERLA agents also showed improved action efficiency compared to vanilla PPO agents, performing fewer actions to achieve the same defensive performance.

\section{Recommendations}

In this research, proof-of-concept PPO-based TERLA agents have been demonstrated in CC4. To mature the approach and explore real-world applicability, we recommend the following pathways for future research:

\begin{itemize}
    \item Apply TERLA to existing ACD agents from Dstl and DARPA research programmes to evaluate whether the extensions can provide similar benefits in independently developed architectures.
    \item Assess TERLA’s ability to address the challenges associated with very large action spaces in large-scale networks, in environments designed to stress test scalability.
    \item Evaluate TERLA in future CAGE Challenge environments, to benchmark performance as challenge fidelity and complexity increase.
    \item Prototype deployment in an emulated network with a real IDS, using TERLA as an advisory capability to investigate human–AI teaming as an initial step towards fully autonomous cyber defence. This would test TERLA’s interoperability with existing security infrastructure and provide insights into how operators might interact with AI-driven recommendations.
\end{itemize}

This additional research will continue to develop knowledge in pursuit of a generalisable cyber defence agent for real-world computer networks.


\begin{thebibliography}{00}
    \bibitem{ncsc1} National Cyber Security Centre, ``The near-term impact of AI on the cyber threat,'' 2024. [Online]. Available: \url{https://www.ncsc.gov.uk/report/impact-of-ai-on-cyber-threat} [Accessed: Feb. 27, 2025]

    \bibitem{ncsc2} National Cyber Security Centre, ``NCSC warns of widening gap between cyber threats and defence capabilities,'' 2024. [Online]. Available: \url{https://www.ncsc.gov.uk/news/ncsc-warns-widening-gap-between-cyber-threats-and-defence-capabilities} [Accessed: Feb. 27, 2025]
    
    \bibitem{army} ``DARPA, DSTL and DRDC form US-UK-Canada AI collaboration,'' Army Technology, 2024. [Online]. Available: \url{https://www.army-technology.com/news/darpa-dstl-and-drdc-form-us-uk-canada-ai-collaboration/} [Accessed: Feb. 27, 2025]
    
    \bibitem{arcd1} Frazer-Nash Consultancy, ``Autonomous resilient cyber defence (ARCD),'' 2025. [Online]. Available: \url{https://www.fnc.co.uk/arcd/} [Accessed: Feb. 27, 2025]

    \bibitem{arcd2} QinetiQ, ``Autonomous Resilient Cyber Defence (ARCD),'' 2025. [Online]. Available: \url{https://www.qinetiq.com/en/capabilities/ai-analytics-and-advanced-computing/autonomous-resilient-cyber-defence} [Accessed: Feb. 27, 2025]
    
    \bibitem{castle} Defense Advanced Research Projects Agency, ``CASTLE: Cyber agents for security testing and learning environments,'' 2025. [Online]. Available: \url{https://www.darpa.mil/research/programs/cyber-agents-for-security-testing-and-learning-environments} [Accessed: Feb. 27, 2025]

    \bibitem{blackhat} I. Miles \textit{et al.}, ``Reinforcement learning for autonomous resilient cyber defence,'' in \textit{Black Hat USA}, 2024. [Online]. Available: \url{https://www.blackhat.com/us-24/briefings/schedule/#reinforcement-learning-for-autonomous-resilient-cyber-defense-39308} [Accessed: Feb. 27, 2025]

    \bibitem{rl} R. Sutton \textit{et al.}, ``Reinforcement Learning: An Introduction,'' MIT Press, edn. 2, 2018, [Online]. Available: \url{http://incompleteideas.net/book/the-book-2nd.html} [Accessed: Oct. 28, 2025]

    \bibitem{environments} G. Palmer \textit{et al.}, ``Deep Reinforcement Learning for Autonomous Cyber Defence: A Survey,'' \textit{arXiv preprint arXiv:2407.17032}, 2021. [Online]. Available: \url{https://arxiv.org/abs/2310.07745} [Accessed: Oct. 28, 2025]

    \bibitem{cyborg} M. Standen \textit{et al.}, ``CybORG: A Gym for the Development of Autonomous Cyber Agents,'' \textit{arXiv preprint arXiv:2108.09118}, 2021. [Online]. Available: \url{https://arxiv.org/abs/2108.09118} [Accessed: Oct. 28, 2025]

    \bibitem{cc4} M. Kiely \textit{et al.}, ``Exploring the Efficacy of Multi-Agent Reinforcement Learning for Autonomous Cyber Defence: A CAGE Challenge 4 Perspective,'' in \textit{Proceedings of the 39th Annual AAAI Conference on Artificial Intelligence}, 2025, [Online]. Available: \url{https://doi.org/10.1609/aaai.v39i28.35158} [Accessed: Jul. 1, 2025]

    \bibitem{gymnasium} M. Towers \textit{et al.}, ``Gymnasium: A standard interface for reinforcement learning environments,'' \textit{arXiv preprint arXiv:2407.17032}, 2024. [Online]. Available: \url{https://arxiv.org/abs/2407.17032} [Accessed: Feb. 27, 2025]

    \bibitem{ppo} J. Schulman \textit{et al.}, ``Proximal policy optimization algorithms,'' \textit{arXiv preprint arXiv:1707.06347}, 2017. [Online]. Available: \url{https://arxiv.org/abs/1707.06347} [Accessed: Feb. 27, 2025]

    \bibitem{challenges} G. Dulac-Arnold \textit{et al.}, ``Challenges of real-world reinforcement learning: Definitions, benchmarks and analysis,'' \textit{Machine Learning}, vol. 110, pp. 2419–-2468, 2021. [Online]. Available: \url{https://doi.org/10.1007/s10994-021-05961-4} [Accessed: Feb. 27, 2025]

    \bibitem{powerflow} Á. López-Cardona \textit{et al.}, ``Proximal policy optimization with graph neural networks for optimal power flow,'' \textit{arXiv preprint arXiv:2212.12470}, 2022. [Online]. Available: \url{https://arxiv.org/abs/2212.12470} [Accessed: Feb. 27, 2025]

    \bibitem{gnn} Jie Zhou \textit{et al.}, ``Graph neural networks: A review of methods and applications,'' \textit{AI Open}, vol. 1, pp. 57--81, 2020. [Online]. Available: \url{https://www.sciencedirect.com/science/article/pii/S2666651021000012} [Accessed: Oct. 28, 2025]

    \bibitem{jobshop1} J. Park \textit{et al.}, ``Learning to schedule job-shop problems: Representation and policy learning using graph neural network and reinforcement learning,'' \textit{arXiv preprint arXiv:2106.01086}, 2021. [Online]. Available: \url{https://arxiv.org/abs/2106.01086} [Accessed: Feb. 27, 2025]

    \bibitem{jobshop2} Z. Yang \textit{et al.}, ``Combining reinforcement learning algorithms with graph neural networks to solve dynamic job shop scheduling problems,'' \textit{Processes}, vol. 11, no. 1571, 2023. [Online]. Available: \url{https://doi.org/10.3390/pr11051571} [Accessed: Feb. 27, 2025]

    \bibitem{rllib} E. Liang \textit{et al.}, ``RLlib: Abstractions for Distributed Reinforcement Learning,'' in \textit{Proceedings of the 35th International Conference on Machine Learning (ICML)}, vol. 80, pp. 3053--3062, 2018. [Online]. Available: \url{http://proceedings.mlr.press/v80/liang18b/liang18b.pdf} [Accessed: Oct. 28, 2025]

    \bibitem{hgt} Z. Hu \textit{et al.}, ``Heterogeneous graph transformer,'' \textit{arXiv preprint arXiv:2003.01332}, 2020. [Online]. Available: \url{https://arxiv.org/abs/2003.01332} [Accessed: Feb. 27, 2025]

\end{thebibliography}
\end{document}